# A first realization of reinforcement learning-based closed-loop EEG–TMS

Dania Humaidan[1,2]†, Jiahua Xu[1,2]†, Jing Chen[1,2], Christoph Zrenner[3,4,5,6], David Emanuel Vetter[1,2], Laura Marzetti[7,8], Paolo Belardinelli[1,2,9], Timo Roine[10], Risto J. Ilmoniemi[10], Gian Luca Romani[8], Ulf Ziemann[1,2]*

[1] Department of Neurology & Stroke, University of Tübingen, Tübingen, Germany
[2] Hertie Institute for Clinical Brain Research, University of Tübingen, Tübingen, Germany
[3] Temerty Centre for Therapeutic Brain Intervention, Centre for Addiction and Mental Health, Toronto, Canada
[4] Department of Psychiatry, Faculty of Medicine, University of Toronto, Toronto, Canada
[5] Institute of Biomedical Engineering, University of Toronto, Toronto, Canada
[6] Institute of Medical Science, University of Toronto, Toronto, Canada
[7] Department of Engineering and Geology, G. d'Annunzio University of Chieti-Pescara, Pescara, Italy
[8] Institute for Advanced Biomedical Technologies, G. d'Annunzio University of Chieti-Pescara, Chieti, Italy
[9] CiMeC, Center for Mind/Brain Sciences, University of Trento, Italy
[10] Department of Neuroscience and Biomedical Engineering, Aalto University School of Science, Espoo, Finland

†Shared first authors

*Corresponding author, ulf.ziemann@uni-tuebingen.de

## Abstract

**Background:** Transcranial magnetic stimulation (TMS) is a powerful tool to investigate neurophysiology of the human brain and treat brain disorders. Traditionally, therapeutic TMS has been applied in a one-size-fits-all approach, disregarding inter- and intraindividual differences. Brain state-dependent EEG–TMS, such as coupling TMS with a pre-specified phase of the sensorimotor µ-rhythm, enables the induction of differential neuroplastic effects depending on the targeted phase. But this approach is still user-dependent as it requires defining an a-priori target phase.

**Objectives:** To present a first realization of a machine-learning-based, closed-loop real-time EEG–TMS setup to identify user-independently the individual µ-rhythm phase associated with high- vs. low-corticospinal excitability states.

**Methods:** We applied EEG-TMS to 25 participants targeting the supplementary motor area–primary motor cortex network and used a reinforcement learning algorithm to identify the µ-rhythm phase associated with high- vs. low corticospinal excitability. We employed linear mixed effects models and Bayesian analysis to determine effects of reinforced learning on corticospinal excitability indexed by motor evoked potential amplitude, and functional connectivity indexed by the imaginary part of resting-state EEG coherence.

**Results:** Reinforcement learning effectively identified the µ-rhythm phase associated with high- vs. low-excitability states, and their repetitive stimulation resulted in long-term increases vs. decreases in functional connectivity in the stimulated sensorimotor network.

**Conclusions:** We demonstrated for the first time the feasibility of closed-loop EEG-TMS in humans, a critical step towards individualized treatment of brain disorders.

# Introduction

Transcranial magnetic stimulation (TMS) is a well-established non-invasive tool to probe and modulate brain activity [1]. Traditionally, TMS has been applied in an open-loop manner—meaning that stimulation parameters such as intensity, target location and timing are pre-determined by the user, without taking into account the individual's changing brain state during a stimulation session. This one-size-fits-all brain state-independent approach inherently disregards inter- and intraindividual variability in brain dynamics.

Brain state-dependent stimulation uses ongoing EEG activity to guide the timing of stimulation [2]. Recently, it was shown that single TMS pulses that are delivered at the trough or ascending phase of the sensorimotor µ-rhythm lead to larger motor evoked potential (MEP) amplitudes than stimulation at other phases [3-8]. Repetitive stimulation at these high- vs. low corticospinal excitability states induced long-term potentiation (LTP)- or long-term depression-like plasticity, respectively [3, 9, 10].

However, current brain state-dependent stimulation protocols remain open-loop, user-dependent and non-adaptive. The stimulation is still triggered based on pre-defined EEG features, without adapting parameters in response to the actual outcome of stimulation. Thus, the system lacks the capacity to learn from feedback and find stimulation parameter values to optimize performance. Closed-loop TMS refers to stimulation protocols that control the expression of a monitored brain state over the course of stimulation by taking the ongoing monitoring information into account for adaptation of the stimulation protocol [2, 11-13].

In this study, we present a first proof-of-concept experiment implementing a reinforcement-learning-based closed-loop real-time EEG–TMS system in healthy humans, where targeting of the phase of the ongoing sensorimotor µ-rhythm is continually optimized over the course of a 400-trial learning period to modify corticospinal excitability as indexed by changes in MEP amplitude. We targeted the supplementary motor area (SMA-proper) to primary motor cortex (M1) network in one hemisphere, which is relevant for movement initiation and motor control [14], and motor recovery after motor stroke [15, 16]. Our findings demonstrate differential excitability and long-term connectivity changes in the SMA-M1 network, depending on the learned instruction to increase or decrease its excitability. The technology may represent an important step towards individualized therapeutic TMS.

## Materials and Methods

**Participants**

The study involved healthy, right-handed volunteers as assessed using the Edinburgh Inventory [17]. Eligibility criteria included 18–50 years of age and the capacity to provide informed consent. Exclusion criteria encompassed a history or presence of psychiatric or neurological disorders, use of drugs affecting the central nervous system, history of alcohol or illicit drug use disorder, current pregnancy, and any contraindications against TMS [18] or MRI. A total of 25 participants (13 female) with an age of 29.3±4.1 years were included in the study. All participants provided written informed consent prior to enrollment. The study received approval from the ethics committee of the medical faculty of the University of Tübingen (525/2021BO2) and adhered to the most recent version of the Declaration of Helsinki.

**Experimental set-up**

The experiment took place in a quiet room with participants seated in a reclined chair. We acquired magnetic resonance imaging (MRI) T1-weighted images using a 3T Siemens PRISMA scanner to accurately position the two TMS coils over left SMA-proper and the hand representation of left M1, relative to each participant's brain anatomy via a neuronavigation system (Localite GmbH, Sankt Augustin, Germany). In addition, the MRI data supported the EEG forward model and source reconstruction (see below).

**M1 stimulation**

Focal TMS was delivered to the hand area of the motor-dominant left M1 using a liquid-cooled butterfly-shaped coil (Cool-B35 HO, MagVenture A/S, Denmark) connected to a MagVenture MagPro R30 stimulator with a biphasic current waveform. The coil was positioned tangentially to the scalp and angled 45° away from the midline to induce a posterior–lateral to anterior–medial (M1–AM) current by the second phase, preferentially activating the corticospinal system transsynaptically via horizontal corticocortical pathways [19, 20]. The optimal coil position for eliciting an MEP in the first dorsal interosseus (FDI) muscle of the right hand was identified as the location where slightly suprathreshold stimulation consistently produced the largest MEPs [21]. The stimulation intensity was then adjusted to elicit MEPs of on average 1 mV in peak-to-peak amplitude. Furthermore, AMT was determined as the minimum stimulation intensity necessary to elicit an MEP >200 µV in the average of 5 trials in the slightly (10–20% of maximum voluntary contraction) isometrically activated FDI [21].

**SMA stimulation**

The left SMA-proper was stimulated with an identical coil as used for M1 stimulation, paired with a MagVenture MagPro X100 stimulator equipped with MagOption and delivering a monophasic current waveform (MagVenture A/S, Denmark). The SMA coil was positioned at the midline, no more than 4 cm anterior to the vertex (i.e., electrode position Cz based on the 10–20 International EEG System), targeting the SMA-proper [22, 23] (see also Fig. S5 in Supplementary Materials). The coil orientation was adjusted to induce a current flow from medial to lateral, and the stimulation intensity was set to 140% AMT [22]. The target position of both coils was monitored against individual brain anatomy throughout the experimental session using the TMS navigation system (Localite).

**EEG recording and pre-processing**

High-density EEG was recorded with a TMS-compatible system (NeurOne, Bittium, Kuopio, Finland) with a sampling rate of 5 kHz, using a TMS-compatible 64-channel Ag/AgCl sintered ring electrode cap (EasyCap GmbH, Germany). Impedance was maintained below 5 kΩ throughout the experimental session. Surface EMG was recorded with bipolar adhesive hydrogel electrodes (Kendall, Covidien) placed over the right FDI at 5 kHz sampling rate. The EEG electrode positions relative to the individualized heads in 3D space were recorded by the neuronavigation system (TMS Navigator, Localite) for source analysis. 5 min of resting-state EEG was recorded at Baseline and 30 min after reinforced learning (RL)-guided training (Post30) (Fig. 1). During the recordings, participants were instructed to remain relaxed with their eyes open and to fixate a cross displayed on a screen 1.5 m in front of them. EEG signals were analyzed using MATLAB version 2024 and the FieldTrip toolbox [24]. Bad channels were first visually inspected and removed (number of removed channels, INCREASE condition: 3.4±2.1, DECREASE condition: 3.4±2.3, RANDOM condition: 4.2±2.5; for definition of conditions, see below). A digital 1–99-Hz bandpass filter and a 50-Hz notch filter were then applied. The data were down-sampled to 1000 Hz and re-referenced using the common average reference method. Removed channels were interpolated with the 'weighted' method [25]. The 5-min

EEG recordings were segmented into 2-s epochs without overlap. Eyeblinks and cardiac signal components were removed using independent component analysis. Signal-to-noise ratio (SNR) was defined as the difference between peak spectral amplitude and fitted noise at that frequency, on the log scale, in units of dB.

**Source construction and functional connectivity estimation**

The head model was constructed from the individual T1-weighted anatomical MRI data. The forward model was calculated using the symmetric boundary element method [26], incorporating conductivities for the scalp, skull, and brain set at 0.33, 0.0042, and 0.33 S/m, respectively [27], whereas the inverse model was generated with a beamforming approach based on the partial canonical correlation method [28], where the spatial filter was calculated by the Dynamical Imaging of Coherent Sources for each location in the source model [29]. Functional connectivity (FC) was estimated using the imaginary part of coherence, which measures to what extent two brain regions synchronize to each other [30]. This method is highly effective in mitigating false connectivity caused by volume conduction, making it well-suited for analyzing FC in resting-state EEG data [31]. The Volume-of-Interest atlas used in this study was the Glasser MMP 1.0 Cortical Atlas (2016) [32], which contains 360 defined brain regions. For this study, we mainly investigated seed-based FC between the two stimulated areas, left SMA-proper, encompassing areas 6ma and 6mp, and left M1. Area 6ma is the medial agranular premotor area (rostral SMA-proper) and area 6mp the medial premotor area (caudal SMA-proper). These areas were included due to their well-established roles in motor planning and execution [33], and their dense anatomical connectivity with M1 [34]. In addition, we extended the analysis to include other nodes of the canonical sensorimotor network, i.e., SMA-proper and M1 on the right hemisphere and the premotor cortex (PM) and primary somatosensory cortex (S1) of both hemispheres.

**Experimental design**

Following [12], we targeted the SMA–M1 motor network in a conditioning–test-stimulus repetitive TMS (rTMS) protocol, applying the conditioning stimulus over the SMA-proper and the test stimulus over the hand area of M1, with an intertrial interval (ITI) of 2–3 s and an interstimulus interval (ISI) of 6 ms. We triggered TMS by targeting different phases of the endogenous sensorimotor μ-rhythm, aiming to achieve a higher paired-pulse MEP (ppMEP) amplitude in the INCREASE condition, or a lower ppMEP amplitude in the DECREASE condition. The algorithm could choose among eight discrete phase bins that were centered from $[-\pi, \pi]$ as follows: $[-\frac{7}{8}\pi, -\frac{5}{8}\pi, -\frac{3}{8}\pi, -\frac{1}{8}\pi, \frac{1}{8}\pi, \frac{3}{8}\pi, \frac{5}{8}\pi, \frac{7}{8}\pi]$.

RTMS was performed in a real-time, closed-loop manner, such that we used an RL framework (Reinforcement Learning Toolbox, MATLAB R2021a), according to the method of Deep-Q Learning (DQN) [35], to find the optimal phase bin to achieve a higher (or lower) ppMEP amplitude. The algorithm (hereafter called the agent) learns from pairs of [Action, Reward]. Each action the algorithm takes (i.e., selecting a phase to target by TMS in the next trial, starting by a random one in the first trial) results in a reward based on the effect of this action, i.e., the ppMEP amplitude in this trial. The Brain Oscillation State Sensor (BOSS) device (sync2brain, Tübingen, Germany) then forward predicts this phase and triggers a pulse at the time of the predicted target phase [3]. Throughout training, the agent learns to select phases that result in higher ppMEP amplitude in the INCREASE condition (or lower ppMEP amplitude in the DECREASE condition), as this leads to higher rewards. Conversely, phases resulting in lower (or higher) ppMEP amplitude are avoided, as they are associated with lower rewards. We defined the phase most used by the agent during the training as the optimal phase. This phase was tested in the post-training evaluations (Fig. 1). The agent was trained for 10 epochs, with each epoch including 40 steps (trials), resulting in a total of 400 trials during training. In the INCREASE condition, we aimed to achieve an increase in the ppMEP amplitude equal to 1.5 times the Baseline value. Accordingly, we defined the reward of the agent as follows:

$$Reward = Current\ average\ ppMEP - 1.5 * Baseline\ average\ ppMEP$$

In the DECREASE condition, we aimed to achieve a decrease in the ppMEP amplitude equal to 0.7 the Baseline value, and the reward of the agent was defined as follows:

$$Reward = 0.7 * Baseline\ average\ ppMEP - Current\ average\ ppMEP$$

The RL parameter values are listed in Table S1 in Supplementary Materials. Fig. 1 illustrates the experimental design. Following preparation, the experimental session started with a Baseline assessment, consisting of 100 paired pulses and 100 single pulses applied at random phase bins in a pseudo-randomized order. Each phase bin is targeted 12-13 times. This Baseline measurement served as the reference for post-training comparisons. Then, the RL training was run. In addition to the INCREASE and DECREASE conditions, participants also underwent a RANDOM condition, in which no RL agent was implemented. Instead of training, stimulation was delivered to the eight phase bins uniform-randomly. The order of experimental sessions across the three conditions was randomized, and consecutive sessions in a given participant were separated by at least one week to avoid carryover effects. Participants were blinded to the stimulation condition; the experimenter and data analyst were not. Immediately after RL training, the first evaluation was conducted, consisting of two subsections. The Post_optimal subsection included 50 paired pulses and 50 single pulses targeting the learned phase, while the Post_random subsection included 50 paired pulses and 50 single pulses delivered at random phase bins in a pseudo-randomized order. These two subsections were tested in a randomized order to prevent order effects. After a 30-min interval, resting-state EEG (rsEEG) was recorded for 5 min. Finally, a second evaluation, identical to the first one (subsections Post30_optimal, Post30_random), was conducted to assess the persistence of learning effects (cf. Fig. 1).

**Statistical Analysis**

We used linear mixed-effects models (LMMs) for the statistical analyses to examine the effects of time (Baseline, Post, Post30), condition (INCREASE, DECREASE) and their interaction on log10-transformed MEP amplitude data to ensure normality (assessed by Kolmogorov–Smirnov tests). A random intercept was included for each participant to account for inter-subject variability. The model formulation was:

$$MEP\ amplitude \sim time * condition + (1|subject)$$

Statistical significance was tested using Wald z-tests; $p<0.05$ indicated statistical significance. Furthermore, 95% confidence intervals (CIs) were reported to quantify parameter uncertainty and present the range of plausibility values for the true effect size. If a CI did not include zero, the effect was considered significant. LMMs were also used to analyze changes in functional connectivity Post30 vs. Baseline. To confirm the results obtained from the LMMs, we implemented a Bayesian mixed-effects model with the same structure. We applied weakly informative priors to all parameters to allow data-driven inference and performed Markov Chain Monte Carlo (MCMC) sampling using the No-U-Turn Sampler (NUTS) with 4 chains and 2,000 posterior draws per chain. Model estimates were summarized using 94% Highest Density Intervals (HDI), providing an intuitive range of the most credible parameter values. If the HDI did not include zero, the effect was considered credible. Furthermore, to quantify the within-condition effects on TIME on MEP amplitude, we examined the posterior distributions of the Post-Pre difference for each condition. To meet the assumptions of normality and homoscedasticity, a log10 transformation was applied to the raw MEP data. Normality was checked using the Kolmogorov–Smirnov test. All statistical analyses were conducted using Python (v3.12). The following packages were used: Statsmodels (v0.14.0) for LMMs, Bambi (v0.12.0) for Bayesian mixed-effects modeling, ArviZ (v0.16.0) for Bayesian inference diagnostics and posterior visualization and Matplotlib (v3.7.2) and Seaborn (v0.12.2) for data visualization.

## Results

### Data set retained for analysis

Rigorous data quality checks resulted in exclusion of 8, 9 and 1 sessions of the INCREASE DECREASE, and RANDOM conditions, respectively, due to low SNR <4 dB of the µ-rhythm in the resting-state EEG at Baseline (Fig. 1) [36]. We retained 29, 22 and 19 sessions of the INCREASE, DECREASE and RANDOM conditions, respectively. In the retained data sets, <1 % of trials were discarded from MEP analysis due to preinnervation in the target muscle, defined as a root mean square EMG signal >200 µV in the period 100 ms to 10 ms prior to TMS.

Importantly, we did not find any statistically significant difference between the INCREASE and DECREASE sessions for SNR ($p$=0.78) (Figs. S1A, C) or phase targeting accuracy in the Baseline resting-state EEG ($p$=1.0) (Figs. S1B, D).

### Performance of the reinforced learning agent

We assumed the existence of a unique optimal phase of the µ-rhythm in a given individual and over the course of the session to achieve either large (in the INCREASE condition) or small ppMEP amplitudes (in the DECREASE condition), respectively. The learnt phase was increasingly often targeted over the 10 learning epochs indicating the occurrence of learning by the agent (Fig. 2). There was no difference in this learning performance between the INCREASE and DECREASE conditions (Wilcoxon signed rank test, $p$>0.05).

### Paired-pulse MEP amplitudes before and after reinforced learning

Subjects participated in up to 5 sessions. Ten subjects participated twice and one subject three times in the INCREASE condition, and 3 subjects participated twice in the DECREASE condition. We performed quantitative analysis of intraindividual ppMEP amplitude variability by calculating the coefficient of variation of the stimulation-induced ppMEP change across repeated sessions. This showed high values (INCREASE condition = 45%, DECREASE condition = 66%), indicating inconsistent effects of stimulation across sessions for the same subject. Furthermore, the significance of the subject-level random effect was checked by performing a Likelihood Ratio Test, which compared the full model with the random effect of the subject and the reduced model without this effect. The results showed negligible variance attributable to the subject-level random intercept (Likelihood Ratio Test = –9.861, df=2, *p*=1.0), and the random effect variance (0.021) was small relative to the residual variance (0.048), suggesting that the variability across sessions within subjects was as large as variability across subjects. Thus, we treated each session as an independent observation to avoid having intra-subject variability conflating with the effects of interest.

LMM statistics demonstrated that the mean ppMEP amplitude in the INCREASE condition was significantly larger at Post compared to Baseline (*p*=0.009, 95% confidence interval (CI) = [0.035, 0.245]) when targeting the learnt optimal phase after training (Post_optimal) (Fig. 3A), while this was not the case when a random phase was targeted (Post) (*p*=0.491, 95% CI = [–0.069, 0.144]) (Fig. 3B). In the DECREASE condition, the mean ppMEP Post_optimal was not different from Baseline (*p*=0.411, 95% CI = [–0.172, 0.070]), but it was significantly lower than the ppMEP Post_optimal in the Increase condition (*p*=0.020, 95% CI = [–0.351, –0.030]), indicating a significant TIME x CONDITION effect (Fig. 3A). Post30_optimal, ppMEP amplitudes showed a significant increase in both the INCREASE (*p*=0.021, 95% CI = [0.018, 0.229]) and DECREASE (*p*=0.023, 95% CI = [0.020, 0.26]) conditions. In the RANDOM condition (i.e., no RL involved), no significant changes were seen in ppMEP amplitude at Post or Post30 compared to Baseline (Fig. 3C).

Very similar results were obtained for the single-pulse MEP (spMEP) amplitude data elicited by single-pulse TMS (spTMS) of M1 (Fig. S2), while there was no change in SMA-to-M1 effective connectivity as expressed by the ppMEP/spMEP amplitude ratio (Fig. S3). Moreover, ppMEP amplitude increased at Post30 compared to Baseline when pooling the data targeting random phase across INCREASE, DECREASE and RANDOM conditions (*p*=0.030, 95% CI = [0.007, 0.147], Fig. S4), suggesting the occurrence of an LTP-like plasticity effect irrespective of condition.

The results of the Bayesian analysis were highly consistent with the LMM findings: ppMEP amplitude significantly increased in the INCREASE condition (mean change Post_optimal = 0.14, standard deviation (sd) = 0.053, 94% highest density interval (HDI): [0.04, 0.25], Fig. 4A), less consistently decreased in the DECREASE condition (mean change Post_optimal = –0.051, sd = 0.061, 94% HDI: [–0.17, 0.07], Fig. 4B), with a significant TIME x CONDITION effect (mean change

Post_optimal Decrease compared with Post_optimal INCREASE: –0.19, sd = 0.083, 94% HDI: [–0.34, –0.04]). These effects were not seen in the Post_random tests, nor in the RANDOM condition. In addition, there was a significant ppMEP increase at Post30_optimal in both conditions (mean change INCREASE Post30_optimal = 0.124, sd = 0.055, 94%, 94% HDI: [0.03, 0.23], mean change DECREASE Post30_optimal 0.139, sd = 0.063, 94% HDI: [0.03, 0.26]).

We reconstructed the mean stimulation sites (i.e., junction of SMA coil) over SMA-proper based on the individual Euclidean distance from Cz in all sessions, using a MNI152 nonlinear symmetric standardized brain template. We found that the stimulation site in 9 sessions was over pre-SMA rather than SMA-proper (Fig. S5A). However, there was no significant difference in Euclidean distances from Cz between Intervention conditions (Fig. S5B) and, crucially, the above reported changes in ppMEP amplitudes when targeting the optimal phase after RL in the INCREASE and DECREASE conditions (Fig. 3A) were maintained when excluding the sessions that targeted pre-SMA (Fig. S8).

**Analysis of optimal phase**

The agent selected an ascending phase of the µ-oscillation in 19/29 (66%) sessions in the INCREASE condition, and a descending phase of the µ-oscillation in 14/22 (64%) sessions in the DECREASE condition (Fig. 5). These distributions are significantly different ($p=0.017$, $\chi^2$-test), and in accordance with expectation from spMEP amplitude modification by phase of the µ-oscillation in previous work testing spTMS of M1 [7, 8].

In those participants that were tested in two sessions of the same condition, we found that in no case the learned phase was identical, suggesting a relevant intrasubject variability across sessions (Fig. S9).

**Changes in functional resting-state EEG brain connectivity**

In the INCREASE condition, we observed a significant FC increase between left SMA and M1 ($p=0.011$). In addition, a significant TIME × CONDITION interaction was detected, explained by differential connectivity changes between the INCREASE and DECREASE conditions post training ($p=0.034$, Fig. 6). No significant connectivity changes Post30 vs. Baseline were found in the DECREASE or RANDOM conditions (Fig. 6), and also not if all conditions were pooled (Fig. S10).

After extending the analysis to include S1 and PM, the results show an FC increase between most connections (10/13) Post30 vs. Baseline in the INCREASE condition, among which the majority (7/10) remains significant after correcting for multiple comparisons. In the DECREASE condition, a trend towards a decrease in FC is observed among most connections. In the RANDOM condition, a few connections (3/13) show a significant FC increase, among which 2/3 remain significant after *p*-value correction for multiple comparisons (Fig. 7).

## Discussion

Our results demonstrate, to the best of our knowledge for the first time, the feasibility of performing closed-loop real-time EEG–TMS, such that the observed effect (ppMEP amplitude) of the action (targeting a phase of the ongoing sensorimotor µ-oscillation) resulted in a participant- and session-specific learning of the phase associated with largest (in the INCREASE condition) or smallest (in the DECREASE condition) ppMEP amplitudes.

This RL-based association of high- (INCREASE condition) vs. low-excitability states (DECREASE condition) with, respectively, the trough/ascending phase vs. the positive peak/descending phase of the µ-rhythm (Fig. 5) is very similar to the findings of testing M1 with spTMS in open-loop offline pre-stimulus µ-rhythm phase sorting [36] or brain state-dependent real-time EEG–TMS experiments [3-7].

In fact, these observed associations between ppMEP amplitudes and µ-rhythm phase may have been driven by effective activation of M1 rather than the SMA-M1 network, as we failed to observe an SMA-to-M1 facilitation with paired-pulse TMS (Fig. S3), in contrast to previous studies [22, 37]. In order to assess possible reasons for this failure, we firstly proved that the conditioning coil in our study was centered over SMA-proper in most of the sessions (Fig. S5A). We secondly checked if the E-field dosage might have been too low in SMA-proper for effective excitation (Figs. S6, S7). The hand representation in SMA-proper has previously been localized to a depth of 7–15 mm from the scalp [38]. At the stimulator intensities we used for the SMA, the E-field strength was on average >50 V/m in SMA-proper at the cortical surface (Fig. S7A), while it was <20 V/m in a depth of 10-15 mm (Fig. S7B). This largely excludes direct excitation of neural elements at that depth, but would be compatible with effective activation of more superficially located neuronal elements, in particular cortico-cortical axons where they are aligned with the local E-field direction [39, 40], e. g., where SMA-proper bends into the interhemispheric sulcus.

Given that we found the agent's preferred phase to vary across sessions in the same participants (Fig. S9), this suggests that the optimal phase varies across sessions, and perhaps even within session [41]. We had originally designed the agent to be able to learn some temporal dynamics of the optimal phase within the session (long short-term memory (LSTM) layer, see Table S1). In practice, however, it became necessary to restrict the agent's policy during evaluation due to limited training data. This likely resulted in suboptimal accuracy of excitability state identification and effect size of plasticity induction.

The learned association of pre-stimulus µ-rhythm phase with corticospinal high- vs. low-excitability states was observed only immediately after the reinforced learning period when specifically testing the learned phase, but not when testing randomly across phases (Fig. 3B) and not when testing the learned phase 30 min after learning (Fig. 3A). The first observation indicates that the learned association was phase-specific, i.e., there was no generalized change in corticospinal excitability immediately after the learning period. The second observation can be explained by the demonstration of an LTP-like increase in MEP amplitude 30 min after learning, when testing at random phase and pooling across all conditions (Fig. 3A, Fig. S4). Four hundred ppTMS were applied to the SMA–M1 network during the learning period, effectively constituting a cortico-cortical paired associative stimulation protocol for plasticity induction [42]. Our findings align with one previous study that has demonstrated an LTP-like effect on MEP amplitude when pairing SMA-proper and M1 stimulation at an interstimulus interval of 6 ms [43]. The induction of this LTP-like plasticity may have overruled the association of µ-rhythm phase with corticospinal excitability.

In contrast to this uniform LTP-like effect on MEP amplitude irrespective of condition, we found a differential effect of condition on FC between the stimulated left-hemispheric SMA-proper and M1 (Fig. 6), and in the bihemispheric canonical sensorimotor network (Fig. 7) 30 min after the learning period. It remains unresolved if this differential effect is caused by stimulation of the SMA–M1 network or if effective stimulation of M1 alone would be sufficient. However, there are arguments in favor of a critical contribution of SMA–M1 network stimulation. Repetitive stimulation of SMA a few milliseconds prior to stimulation of M1 at a time of high excitability of M1 would lead to strengthening of this connection, according to Hebbian plasticity rules, while stimulation at a time of low excitability of M1 would lead to weakening [44, 45]. Accordingly, the FC of the stimulated left SMA-proper with other areas of the canonical sensorimotor network demonstrated the differential changes in the INCREASE vs. DECREASE conditions (Fig. 7). In contrast, the concurrently observed uniform LTP-like increase in corticospinal excitability irrespective of condition 30 min after reinforced learning (Fig. 3A, Fig. S4) likely reflects a postsynaptic effect of repetitive corticocortical paired associative stimulation in M1, in accord with rules of spike-timing-dependent plasticity [42, 46, 47].

In summary, these findings indicate that our closed-loop real-time EEG–TMS algorithm learned to identify phases of the ongoing sensorimotor µ-rhythm that are associated with high- vs. low-excitability states of the corticospinal system. This is a significant advancement in comparison to brain state-dependent stimulation where the state parameters are preset and fixed (e.g., the trough of the ongoing sensorimotor µ-rhythm), under the assumption that they always represent a certain state (e.g., a corticospinal high-excitability state) [3-5, 48]. The present findings show that this is not the case for all subjects (Fig. 5), and also not across subsequent sessions in a given individual (Fig. S9). To account for this variability, adaptive closed-loop approaches are critically needed. To what extent adaptive closed-loop brain stimulation will hold the promise to be superior in the induction of long-term plasticity and therapeutic effects compared to open-loop approaches will need to be demonstrated in future studies [13].

While alternative learning approaches, such as Bayesian optimization, can be effective for tuning stimulation parameters in TMS experiments [49, 50], we chose to employ RL because our long-term goal is to extend the optimization framework beyond a single EEG parameter.

This study has several limitations. First, the dataset was imbalanced regarding the number of retained sessions. Second, we observed a lack of SMA-to-M1 facilitation. Ensuring robust ppTMS effective connectivity prior to machine-based learning could improve future experimental designs. Third, we have not tested behavioral consequences of the observed changes in corticospinal excitability and functional connectivity in the canonical sensorimotor network. This should be done in future studies. Fourth, the learning agent deployment during testing was limited. Allowing the agent to employ its full learned policy might reveal multiple viable phase options and potentially improve stimulation outcomes. Finally, our approach allowed the reinforced learning algorithm to choose from only 8 phase bins of the ongoing µ-rhythm. There is now evidence that corticospinal excitability states can be predicted, in addition to the phase of the µ-rhythm by an ample set of other EEG metrics, such as µ-rhythm power [51], the interaction of µ-rhythm power and phase [6, 52], EEG–EEG functional connectivity in the motor network [53, 54], or even complex spatiotemporal EEG signatures identified in purely data-driven machine learning-based approaches [55, 56]. Therefore, algorithms in future experiments may employ learning approaches on less restricted EEG feature sets to enable even more effective individualized neuromodulation.

**Conclusions**

Collectively, this study presents the first realization of an user-independent closed-loop real-time EEG–TMS system in the human motor network to induce differential long-term changes in connectivity, showcasing its potential to advance therapeutic neuromodulation towards an individualized self-optimizing approach. The significant increase in corticospinal excitability and SMA-to-M1 functional connectivity when pairing SMA and M1 stimulation at instances of high M1 excitability highlight promising applications for neurorehabilitation.

**Acknowledgements**

We would like to thank Johanna Metsomaa, Elena Ukharova, Mikael Laine, Roberto Guidotti, Maria Ermolova, Miriam Kirchhoff, Giuseppe Varone, Saeed Makki, Matilda Makkonen, Hanna Pankka and Ida Granö for their valuable assistance with the initial preparations and discussions.

**Funding**

This study is part of the ConnectToBrain project that has received funding from the European Research Council (ERC) under the European Union's Horizon 2020 research and innovation program (Grant agreement No. 810377).

**Author contributions**

Conceptualization: DH, TR, RI, GLR, UZ; Methodology: DH, JX, DV, CZ, PB, LM; Investigation: DH, JX, JC; Supervision: UZ; Writing—original draft: DH, JX; Writing—review & editing: DH, UZ

**Competing interests**

RI has patents and patent applications on TMS technology. All other authors have no conflicts of interest.

**Data and materials availability**

The data that support the findings of the manuscript are available upon request under specified conditions, i.e., EULA (End User License Agreement). The code of the analysis in this study is available at https://github.com/bnplab/mPOP

Figures

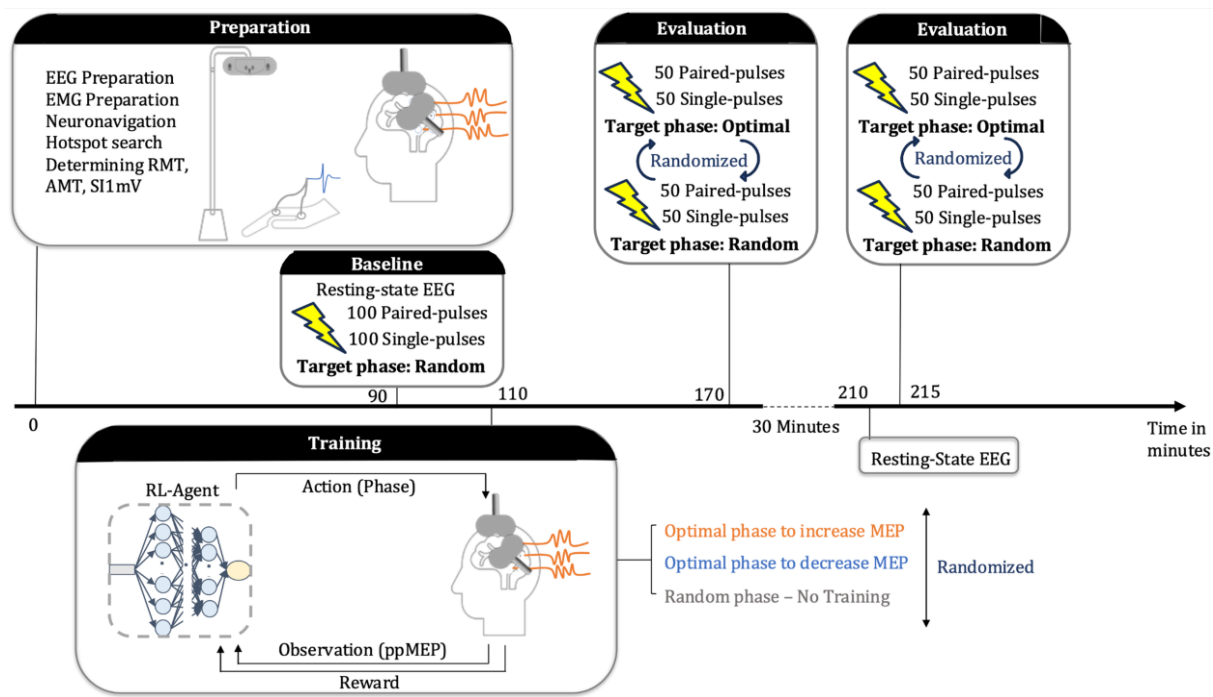

**Fig. 1 | Experimental design of the study.**

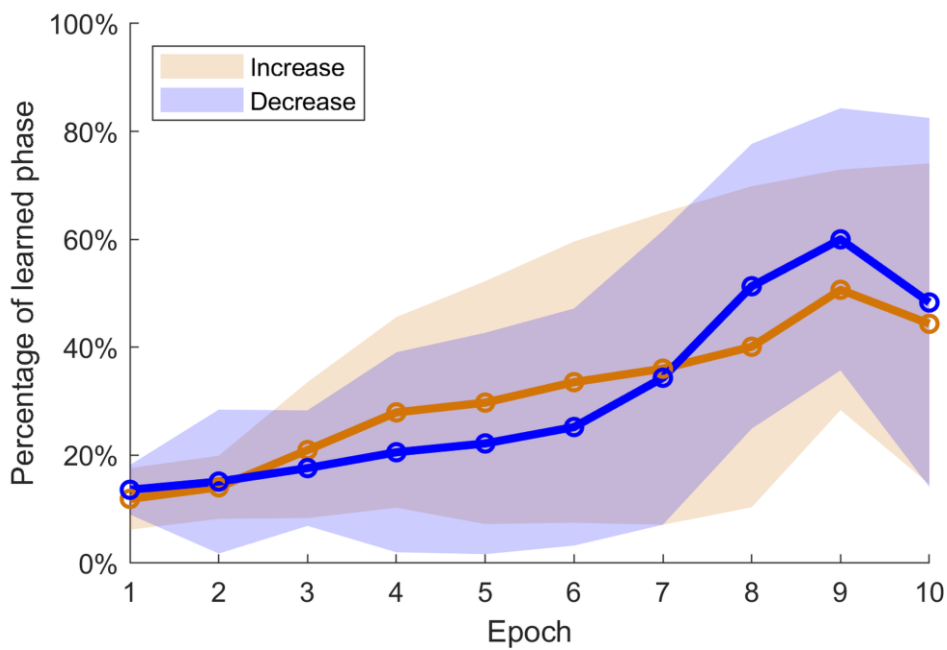

**Fig. 2 | Performance of the reinforced learning agent.** Mean percentage of trials in which the optimal phase identified by the end of training was used during each epoch of training (40 trials each), in the INCREASE (red, $n$=29 sessions) and DECREASE condition (blue, $n$=22 sessions). Shaded areas represent ±1sd of the percentage of trials using the learnt phase.

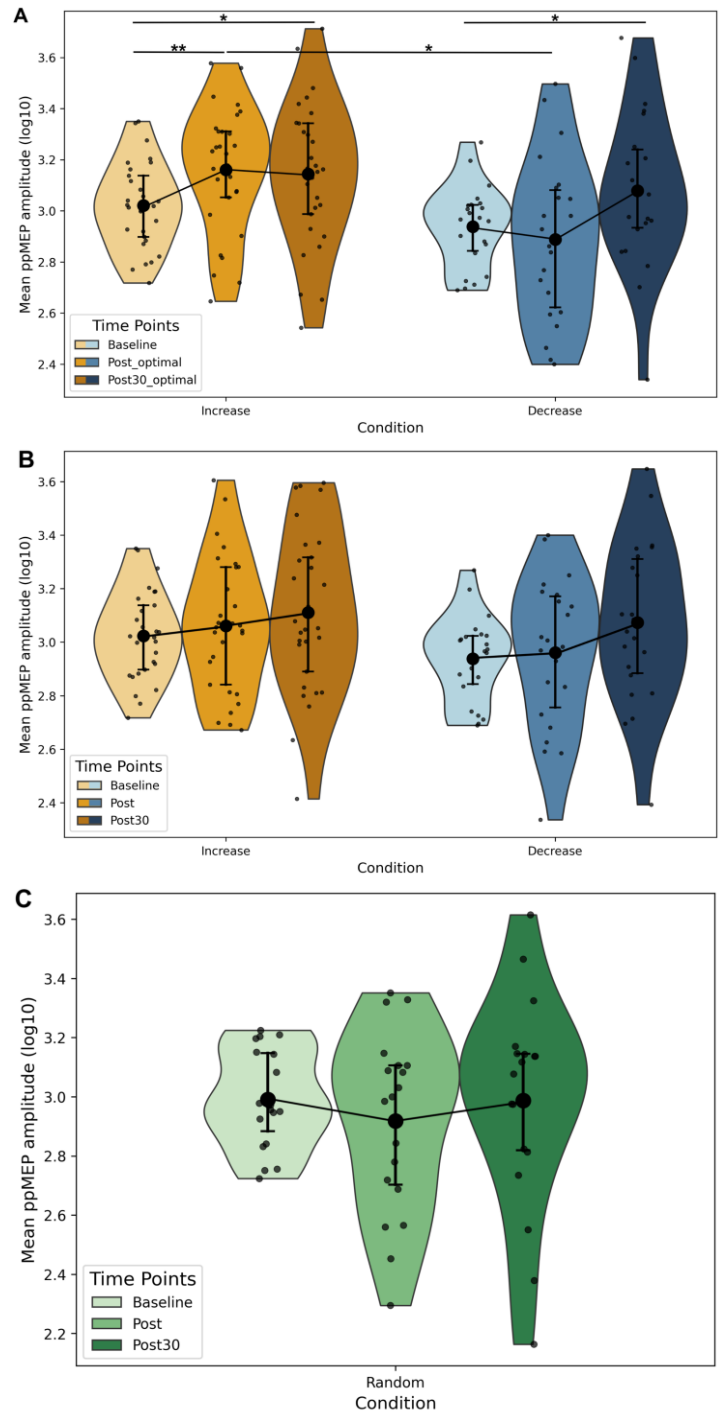

**Fig. 3 | Effects of reinforced learning on paired-pulse motor evoked potential (ppMEP) amplitude. A** Change in ppMEP amplitudes when targeting the optimal μ-rhythm phase immediately (Post_optimal) and 30 min (Post30_optimal) after reinforced learning in the INCREASE and DECREASE conditions. **B** Change in ppMEP amplitudes when targeting random phase immediately (Post) and 30 min (Post30) after reinforced learning in the INCREASE and DECREASE conditions. **C** Change in ppMEP amplitudes in the RANDOM condition, i.e., without reinforced learning involved. Baseline ppMEP amplitudes were tested always at random phase prior to the learning period. Violin plots, with large black dots representing mean ppMEP amplitudes, vertical bars showing the 25th and 75th percentiles, and small black dots indicating individual data. * $p<0.05$; ** $p<0.01$.

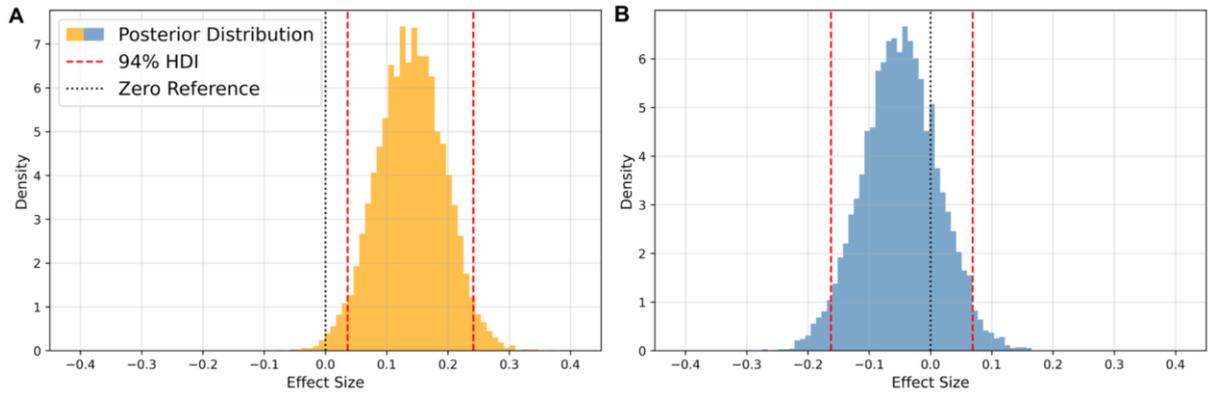

**Fig. 4 | Posterior distribution of the effect size of time on ppMEP amplitude**, showing the estimated effect size for Post_optimal relative to Baseline. **A** In the INCREASE condition; **B** in the DECREASE condition. The x-axis represents the effect size, while the y-axis represents the posterior density. The red dashed lines indicate the 94% Highest Density Interval (HDI), which contains the most credible range of values for the effect. The black dotted line at zero represents no effect. If the 94% HDI excludes zero, the effect is considered credible.

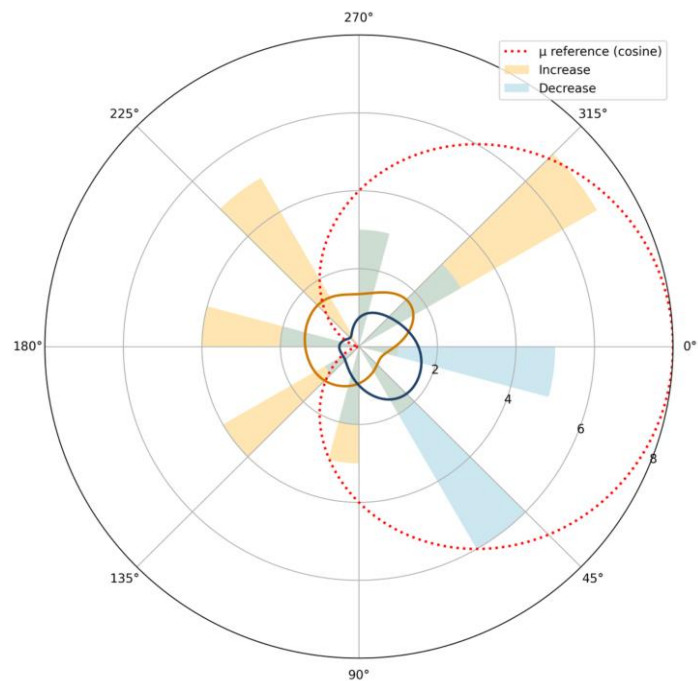

**Fig. 5 | Distribution of phases of the μ-oscillation associated with largest or smallest ppMEP amplitudes identified by reinforced learning in the INCREASE (orange) and DECREASE (blue) conditions.** The dotted red cosine shows the μ-oscillation with 180° indicating the trough and 0°/360° the peak. The orange and blue curves are the kernel distribution estimations (KDE) in the INCREASE and DECREASE conditions, respectively. The corresponding Pearson correlation coefficient ($r$, $p$-values) for the INCREASE and DECREASE conditions were (-0.389, <0.0001) and (0.880, <0.0001), respectively. Rayleigh's test indicated significant clustering only in the DECREASE condition ($p$=0.015), while the INCREASE condition showed a broader distribution. Circular–circular correlations with a cosine-phase template were nonsignificant in both conditions, consistent with the phases not being linearly related to the template on a sample-by-sample level. Circles indicate steps of $n$=2 observations.

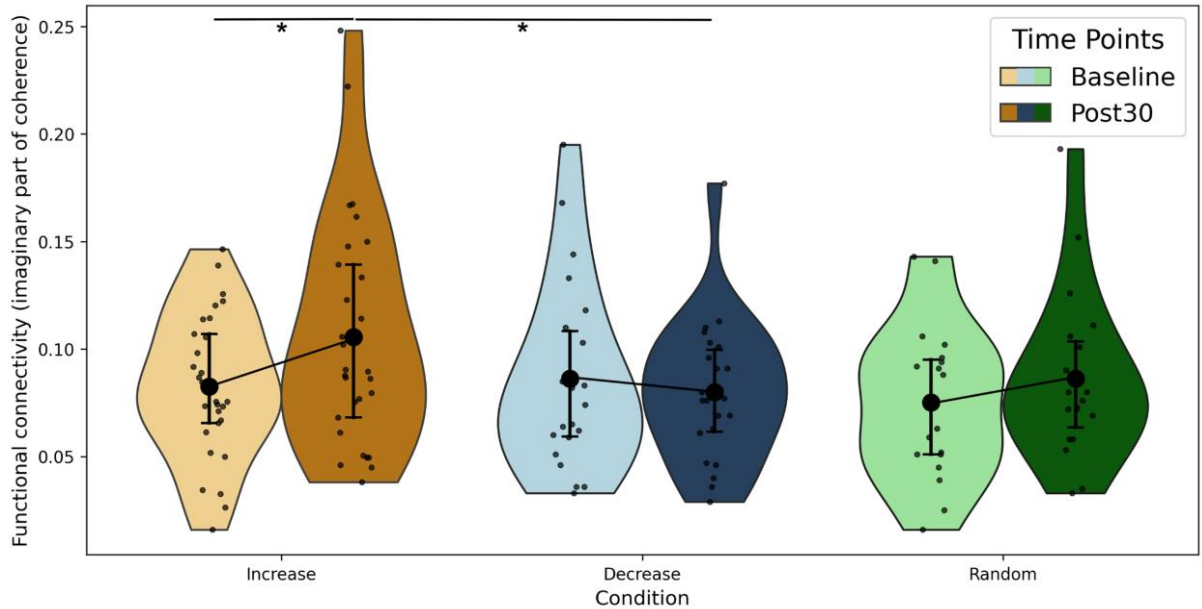

**Fig. 6│ Effects of reinforced learning on resting-state EEG functional connectivity between the stimulated left SMA and left M1 in the different conditions (INCREASE, DECREASE, RANDOM) after reinforced learning (Post30) compared to Baseline.** Mean changes between Post30 and Baseline: INCREASE: 0.023±0.009; DECREASE: –0.006±0.010; RANDOM: 0.011±0.011. Violin plots, with large black dots representing means of the imaginary part of coherence, vertical bars showing the 25th and 75th percentiles, and small dots indicating individual data. * *p*<0.05

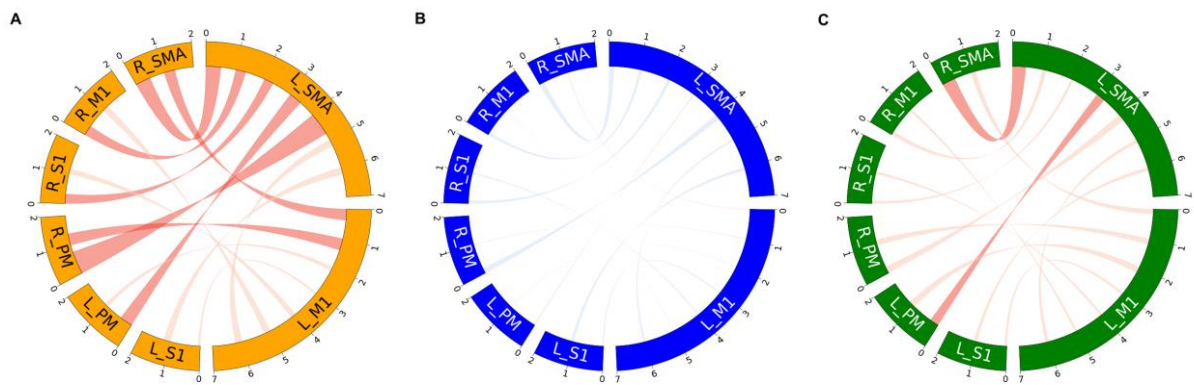

**Fig. 7│ Effects of reinforced learning resting-state EEG functional connectivity of the stimulated left SMA and left M1 with other nodes in the canonical sensorimotor network** in the different conditions (**A** INCREASE, **B** DECREASE, **C** RANDOM) after reinforced learning (Post30) compared to Baseline, corrected for multiple comparisons. Dark red connections indicate a significant increase in connectivity, light red/blue connections indicate an insignificant increase/decrease in connectivity. M1: Primary motor cortex, SMA: Supplementary motor area, PM: Premotor cortex, S1: Primary somatosensory cortex, R: right, L: left.

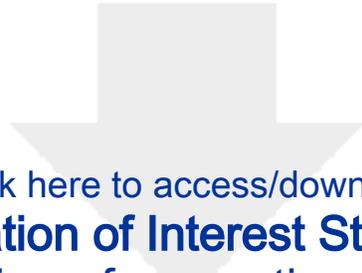

Click here to access/download
**Declaration of Interest Statement**
BRS declaration-of-competing-interests.docx

Supplementary Materials

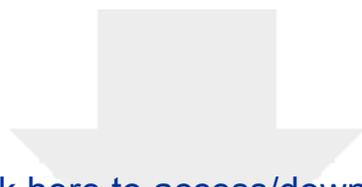
Click here to access/download
**Supplementary Item**
Supplementary_Material.docx